\documentclass{article}
\usepackage{authblk}
\usepackage{graphicx}
\usepackage{subcaption}
\usepackage{url}            
\usepackage{listings}
\usepackage{tikz}
\usetikzlibrary{positioning}
\usepackage{pgfplots}
\pgfplotsset{width=8cm,compat=1.16}
\usepackage{standalone}
\usepackage{verbatimbox}
\usepackage{hyperref}
\usepackage{microtype}
\usepackage{graphicx}
\usepackage{url}
\usepackage{amssymb}
\usepackage{booktabs}
\usepackage{xcolor}
\usepackage{xspace}

\newcommand{\eps}{\epsilon}
\newcommand{\modelmonitor}{{\tt Model Monitor}\xspace}

\title{Amazon SageMaker Model Monitor: A System for Real-Time Insights into Deployed Machine Learning Models}

\author[1]{David Nigenda}
\author[1]{Zohar Karnin}
\author[1]{Muhammad Bilal Zafar}
\author[1]{Raghu Ramesha}
\author[1]{Alan Tan}
\author[1]{Michele Donini}
\author[2]{Krishnaram Kenthapadi\thanks{Work done while at Amazon AWS AI}}
\affil[ ]{{\small{\{nigenda, zkarnin, zafamuh, ragmesh, alantan, donini\}@amazon.com}}}
\affil[ ]{{\small{krishnaram@fiddler.ai}}}
\affil[1]{Amazon AWS AI}
\affil[2]{Fiddler AI}

\begin{document}

\maketitle

\begin{abstract}
With the increasing adoption of machine learning (ML) models and systems in high-stakes settings across different industries, guaranteeing a model's performance after deployment has become crucial. Monitoring models in production is a critical aspect of ensuring their continued performance and reliability. We present Amazon SageMaker Model Monitor, a fully managed service that continuously monitors the quality of machine learning models hosted on Amazon SageMaker. Our system automatically detects data, concept, bias, and feature attribution drift in models in real-time and provides alerts so that model owners can take corrective actions and thereby maintain high quality models. We describe the key requirements obtained from customers, system design and architecture, and methodology for detecting different types of drift. Further, we provide quantitative evaluations followed by use cases, insights, and lessons learned from more than two years of production deployment.

\end{abstract}

\section{Introduction}\label{sec:intro}
Machine learning (ML) models are becoming more common and playing critical roles in businesses worldwide. However, over time, their performance can degrade due to unexpected or unintended changes in the data collection process, or due to changing real-world conditions causing the training data, and therefore the model, to become stale \cite{sculley2015hidden}.
Models need to be monitored to ensure that they continue to meet predictive quality requirements; this monitoring is a vital part of the MLOps lifecycle (i.e, the set of practices for deploying and maintaining ML models in production reliably and efficiently). While retraining models too frequently can be expensive, not retraining often enough can result in sub-optimal predictions. Continuous monitoring of production models can be used to identify the right time and frequency to retrain the model, and more broadly address the following key challenges:\\ 

{\noindent \em Data drift}: Data distribution in the real world is rarely static. 
The quality of the model predictions may gradually degrade over time as a result of a drift in the production data distribution, compared to the training data distribution~\cite{breck2019data}.\\

{\noindent \em Changes in the environment and relationships between variables}: Relationships between input and target variables can evolve over time due to changes in the environment. 
This phenomenon, known as concept drift~\cite{gama2014survey, tsymbal2004problem}, can lead to suboptimal model performance.
For instance, the dependence of the target variable on features changes rapidly in adaptive and adversarial settings such as when monitoring for fraud, spam, and phishing: this necessitates frequent feature engineering and model retraining.\\

{\noindent \em Operational changes}: Operational changes in the data pipeline (e.g., switching feature units from feet to inches) that can lead to model generating erroneous outputs. In contrast to traditional software or application monitoring, operational issues with ML models can often go unnoticed.
Building an effective monitoring system is challenging as it requires solving two different tasks: (i) building tools to store prediction-related data securely, and at scale, which requires expertise in distributed systems, software engineering, and storage systems; (ii) implementing various statistical techniques to analyze this data and to evaluate the quality of the model, which requires expertise in data science and statistics. 
Data scientists and ML engineers often lack expertise in
(i), (ii), or both, 
or do not have the resources to build a monitoring system at scale.\\

In this paper, we present Amazon SageMaker \modelmonitor,\footnote{\scriptsize \url{https://aws.amazon.com/sagemaker/model-monitor}}
a fully managed service that continuously monitors the quality of ML models hosted on Amazon SageMaker. \modelmonitor eliminates the need for developers to build any tooling to monitor models in production and detect when corrective actions need to be taken once the model is deployed.
It does this by, at a regular frequency, automatically analyzing the collected data based on user-provided rules (usually the user is a data scientist) to determine if there are any rule violations. 
Developers can also use \modelmonitor’s built-in statistical rules to analyze tabular data, automatically detect common issues in production, and receive alerts.

\modelmonitor provides these capabilities as pre-built monitoring solutions that do not require coding, while also providing the flexibility for the user to perform custom analyses. Specifically, \modelmonitor provides the following types of monitoring:\\

{\noindent \bf Monitor Data Quality}: It automatically captures and monitors the input given to ML models, uses rules and statistical tests to detect both data drift and unexpected upstream data changes, and raises alerts when necessary.\\

{\noindent \bf Monitor Model Quality}: It monitors the performance of a model by capturing its predictions and comparing them with the actual ground truth labels that the model attempts to predict. To do this, it merges data that is captured from real-time inference with actual labels stored in 
Amazon S3,\footnote{\scriptsize \url{https://aws.amazon.com/s3/}} a storage service, 
and then compares them using metrics specific to the ML problem type.
These could be either pre-built standard metrics such as mean square error (MSE) and accuracy, or 
custom user-defined metrics.\\

{\noindent \bf Monitor Bias Drift for Models in Production}: Bias has appeared as a concerning potential issue in ML systems. 
A computer system may be considered biased if it discriminates against certain societal groups \cite{barocas2016big}. 
Although the initial data or model may not have been biased, data or concept drift (among other factors) may cause the model outcomes to become biased over time. 
For example, a substantial change in home buyer demographics could cause a home loan application model to become biased if certain populations were not present in the original training data.
\modelmonitor/ supports monitoring of several bias metrics \cite{das2021fairness} computed over data captured from the model's inputs and outputs during inference, and issues alerts when a metric exceeds a specified threshold.\\

{\noindent \bf Monitor Feature Attribution Drift for Models in Production}: Feature attribution drift means that a feature that was once not influential in the prediction process became influential or vice versa. This can be an early indication of degradation in the model quality \cite{google_drift}. 
\modelmonitor supports monitoring of drift in attribution for individual features and changes in the relative importance of features, offering metrics measuring attribution change and the option to issue an alert once a metric deviates from a certain threshold.

The rest of the paper is organized as follows. We first discuss the role of model monitoring in the ML life-cycle and briefly describe Amazon SageMaker in \S\ref{lifecycle}. We present the design considerations and technical architecture of \modelmonitor as well as its integration in the ML pipeline in \S\ref{design}. We then describe the methodology for data drift detection and application for detecting drift in model bias metrics and feature attributions in \S\ref{sec:methodology}. We provide quantitative evaluations in \S\ref{sec:quantitative}, followed by a case study, common production use cases, and deployment insights in  \S\ref{use-cases}. Finally, we discuss related work in \S\ref{sec:related} and conclude in \S\ref{sec:conclusion}.

\section{Background}\label{lifecycle}
\subsection{Role of model monitoring in ML lifecycle}

A common ML setting is to build and deploy models that produce accurate inferences. Models are commonly trained using historical data: However real-world data received by the model might look different than the training data. For example, a model for forecasting vacation rental demand, trained using data from before a global pandemic, would not make accurate predictions during the pandemic. 
Changes in the relationship between input and target variables (i.e., concept drift) can lead to degradation of model accuracy, and consequently business or customer impact. 
As another example, consider changes in the data collection pipeline where some format changed and is no longer parsed properly, or a numeric value changed scale (e.g., kg to gram). Such events are common in an industry setting and often derail the performance of a deployed model.
To make sure models continue to make accurate and relevant predictions, regular monitoring of the input and inferences needs to be incorporated into the ML lifecycle which can be broadly categorized into the following steps:
\begin{enumerate}
    \item Prepare Data:  Data form the foundation for building an ML model. This stage of the lifecycle focuses on identifying, cleaning, pre-processing, and engineering dataset features. The dataset should be representative of real-world observations the model is expected to predict. Notice that the data wrangling steps - format conversions, joining database tables, etc. - will be run on a regular basis once the model is deployed, and should hence be monitored.
    \item Build, Train, and Tune Model: Algorithms and frameworks are selected based on the data and the ML problem. The model is tested and tuned iteratively until it meets the performance requirements.
    \item Deploy Model: Model starts making predictions based on new real world data. The model is integrated with other systems that are already in place to surface inferences made by it.
    \item Monitor: Monitoring the deployed model is crucial for ensuring its continued performance and reliability. 
    The inputs and outputs of the model are compared against the training dataset and expected values, to validate that the model continues to meet performance requirements. This step should be automated so that drift and data bugs are detected in a timely manner. When an issue is detected, the lifecycle is restarted by either fixing the problem source or preparing and retraining the model using new real world data.
\end{enumerate}

\subsection{Amazon SageMaker}

\begin{figure}[htbp]
\centering
  \includegraphics[width=12cm]{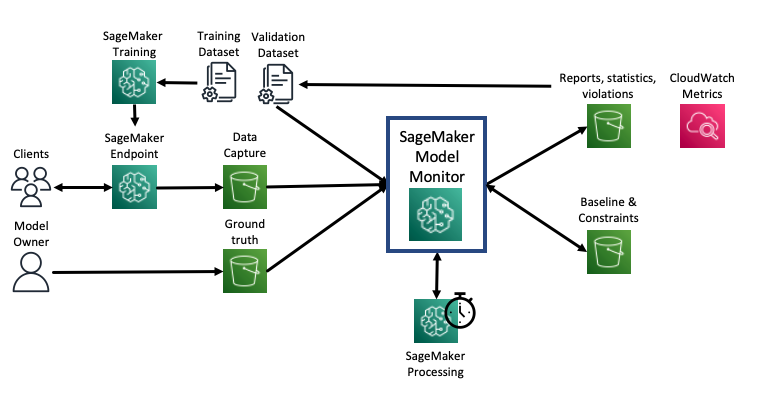}
  \caption{High-level system architecture of \modelmonitor.}
  \label{architectureDiagram}
\end{figure}

Amazon SageMaker\footnote{\scriptsize \url{https://aws.amazon.com/sagemaker}} helps data scientists and developers to prepare, build, train, and deploy high-quality ML models quickly by bringing together a broad set of capabilities purpose-built for ML. 
It offers 
fully managed
tools for every step of ML development, including labeling, data preparation, feature engineering, bias detection, auto-ML, training, tuning, hosting, explainability, monitoring, and workflows.
Once a model is trained, it can be deployed into production with a single click to start generating predictions. The models can be deployed onto auto-scaling instances across multiple availability zones for high redundancy.
\modelmonitor was designed to close the loop (step 4 in \S\ref{lifecycle}) in the ML lifecycle when using SageMaker.

\section{Design and Architecture}\label{design}
\subsection{Key Requirements}
We next summarize key design requirements, gathered from customers and domain experts, that were used to shape \modelmonitor.

{\bf Be accessible:} Model monitoring is a key part of the ML lifecycle and a critical MLOps component: it should be approachable by customers with any level of expertise, both in terms of engineering and ML. There should be minimal amount of code needed to get started, and it should provide sensible defaults where possible. However, it should not preempt more advanced use cases.

{\bf Be scalable:} Frequently used models in production settings can generate large amounts of data. Processing this data should not impact inference; additionally the analysis has to be fast and scale horizontally.
    
{\bf Handle historical data:} Analysis is done not only on the most recent data but on historical data as well. This data, a sample of it, or its summary should be stored in a way that is securely and readily accessible.
    
{\bf Incorporate external sources:} Analysis often requires not only model inputs and outputs but external sources such as labels that may only be known with some delay after inference itself. It should be easy to quickly access the inference traffic data together with its corresponding labels or other external information.

\subsection{\modelmonitor Architecture} \label{sec:components}
\modelmonitor consists of three major components: data analysis, data collection, and scheduling.

\subsubsection{Data Analysis \label{data_analysis}}
The data analysis component strives to achieve these design goals:
(1) scale to any amount of inference data, (2) provide sensible defaults for ease of use, and (3) provide an interface that can be easily and intuitively extended (for instance by scientists or domain experts), e.g.,\ allowing custom metrics.

\modelmonitor performs data analysis by leveraging SageMaker Processing,\footnote{\scriptsize \url{https://docs.aws.amazon.com/sagemaker/latest/dg/processing-job.html}} a scalable cloud platform for ETL workloads. The data analysis component contains the core analysis logic and is packaged as a Docker image for use in SageMaker Processing. Users with different analysis needs have the option of providing a custom
Docker image.
SageMaker Processing runs customer-defined analyses by loading Docker images: for some metrics we turned to proven open-source technologies to drive the core analysis logic, using amongst others, Apache Spark MLlib and Deequ \cite{schelter2018automating}. In some cases, we contributed back to open source projects, e.g., by extending the functionality of Deequ allowing it to track an approximate histogram of the data.
Our use of Apache Spark and S3 for data collection allows us to easily distribute the load amongst instances provisioned by SageMaker Processing, and scale as needed.

Upon startup, the analyzer loads relevant inference data, as well as labeled data (when applicable),
and produces a time-bound analysis. It computes a standard set of metrics by default, which we describe in depth in \S\ref{sec:methodology}. We show a sample generated report in Appendix \ref{app:samplereport}.

We support custom analyses as follows.
For use cases where our default metrics are not a good fit, a custom Docker image can be provided. Note that, in this scenario, the analysis is still run on the data collected through our Data Collection system described in \S\ref{data_collection}. If however, small changes prior to analysis are required, pre-processing (run over each row) and post-processing (run after analysis is done) Python hooks are provided. While our analysis is done in Spark, we chose to provide hooks in Python due to its popularity in the ML community.

\subsubsection{Data Collection \label{data_collection}}
SageMaker Endpoints is a fully-managed real-time model hosting platform. As part of \modelmonitor, we built a data collection (also referred to as data capture) component to (1) capture and collect the inputs and outputs sent to these Endpoints
and (2) a way to join this data with external signals such as labels/annotations.

The first part operates with no code change on the Endpoints as it captures only inputs and outputs, and not intermediate data. This is crucial for ease of use, especially for already existing systems that wish to start using \modelmonitor. See Appendix \ref{app:enabling} for details into this area. Other than ease of use, our main focus for data collection was to avoid impact to inference workloads: extraneous resource usage (CPU, memory, disk, network) must be minimal.

Minimizing resource usage is a well studied problem in logging systems. We approached this problem in two stages: (1) requests and responses are saved to disk in a format that is easy to break up (JSON lines),  and (2) a daemon polls the disk for changes - upon reaching thresholds for size and time between uploads, the daemon initiates the transfer into S3. 
For ease of analysis, we store the data in sub-folders structured as yyyy/mm/dd/hh. Further, we exclusively use UTC timestamps to avoid confusion related to multiple time zones and non-existing hours due to daylight savings time changes.

Stage (2) above is required for tracking performance metrics such as accuracy. To use it, users are responsible for adding an identifier in each inference request, which gets saved in S3 along with the rest of the data. They must then emit the ground truth label or value into S3 in the same hourly sub-folder format we use for inference requests, each with the corresponding identifier. 

\subsubsection{Scheduling \label{scheduler}}

This component controls the orchestration between the inference nodes and other \modelmonitor components. 
In particular, this component allows users to configure the analysis hardware (e.g., size and number of instances) and frequency.

For simplicity, \modelmonitor performs analyses in hourly batches. This was a coarse enough granularity to provide a sensible way to store in S3, but granular enough to provide a useful signal upon analyzing the data. Storing the data in this manner meant, however, that analysis was constrained to run at the top of the hour, if we wanted to analyze the preceding hour's data as soon as possible. Implementing this approach naively would cause imbalance in terms of compute. Our solution was to add a random delay to each job of up to twenty minutes, which spreads jobs more evenly and 
avoids overloading our service at the top of every hour.

Note that it could be possible not to persist the joined data, and instead run the analysis on it and discard it shortly after. However, to allow multiple components to perform analysis on this data - 
our own, users' custom components, and third party monitoring tools
- we found it to be a better design choice to have a separate job for joining the data as well as persisting it in S3. The scheduler runs this ``join'' job before running the analysis that requires the data.

\subsection{Integration with Amazon SageMaker} \label{architecture}

\modelmonitor was built to provide an integrated and seamless experience for customers of SageMaker and Amazon Web Services. As shown in Fig.~\ref{architectureDiagram} and described in the preceding section, \modelmonitor is built to understand and analyze data collected by SageMaker Endpoints. The amount of data to be collected can be tuned taking into account storage and compute costs. In addition, data generated by \modelmonitor are stored in 
Amazon S3 (see Appendix \ref{app:samplereport} for an example),
enabling the export and use of other AWS Services to perform additional analysis on the data. \modelmonitor jobs also emit metrics to Amazon CloudWatch,\footnote{\scriptsize \url{https://aws.amazon.com/cloudwatch/}} a monitoring service, to indicate whether a deviation in quality was identified. Users can set alarms on these metrics to alert operators, who can then take remedial action. For instance, an operator may realize, upon seeing a ``mismatched number of columns alert'' that an upstream service is no longer one-hot encoding a feature. In such a scenario, reverting the deployment may be the most expedient course of action.

\section{Model Monitoring Methodology}\label{sec:methodology}
We discuss the methodology underlying drift detection in \modelmonitor, focusing on the \emph{Data analysis} component described in \S\ref{data_analysis} as well as model bias metrics and feature attributions.

\subsection{Core Methodology: Data Sketches}
A useful concept in recording metrics is that of a data sketch. To explain it, consider a simple example where we have a regression model that outputs a single number and we wish to monitor its mean over time. A sketch is an object that is fed a stream of numbers, in this case the output of the model. It maintains (1) {\bf tot}, the total amount of numbers it observed and (2) {\bf sum}, the sum of numbers observed so far. For every model output observed it \emph{updates} its internal state, meaning it adds 1 to {\bf tot} and the value to {\bf sum}. At any given point one can query this sketch for the current mean, and it will return it by \emph{finalizing} its internal state containing {\bf tot, sum}, in this case by computing {\bf sum/tot}. If we have two machines hosting the same model, both will have an internal state containing their local copies of {\bf tot, sum}, and when querying the overall mean, they will \emph{merge} into a single state by adding their respective {\bf tot, sum}, and \emph{finalize} that state, providing the overall mean. In cases where users wish to query for the mean at different time intervals, say, of granularity up to one hour, we can store a state for every hour, and given a query over an interval of X consecutive hours, \emph{merge} the relevant states, \emph{finalize} the merged state, and return the mean.

This paradigm extends beyond this simple example of calculating a mean, and we use it for many different metric computations. The application is sometimes straightforward for metrics such as mean, standard deviation, and count or proportion of missing items. In other cases it is more challenging or even impossible to obtain an exact measurement. For example, if we wish to maintain a histogram of the observed values, or even just the median, it is known to be impossible to have a precise metric without storing all the observed items. Fortunately, it is almost always the case that an approximation suffices, meaning the users are content with a number which is the p49-p51 rather than the exact median. In this particular case, we can use known algorithms from the literature, such as \cite{karnin2016optimal} providing a sketch supporting the \emph{update, merge, finalize} operations, or its follow-up \cite{cormode2021relative} solving the same problem tuned for tail values, e.g.,\ the 99.9th percentile. For tracking histograms, we have in fact implemented a version of \cite{karnin2016optimal} in the open-source Deequ package. Our implementation provides an interpretable sketch whose internal variables can be serialized in JSON format; we found this to be a useful property for users that wish to run custom code or simply understand the mechanism of this sketch.  

There are many different metrics that can be tracked via sketches. Since we do not (and cannot) provide a sketch for all of these we maintain a sample of data points. These are useful as-is for our users, and can be seen as a `catch-all' sketch for these mentioned metrics, custom metrics, or metrics without a known sketch. One could wonder why anything other than a sample is required at all. The answer is that by computing the metrics on a sample we obtain only an approximate answer. With sketches we can obtain a more accurate answer with less resources (both computation and storage), as in the case of approximate histograms,\footnote{For an $\eps$-approximate approximation \cite{karnin2016optimal} requires roughly $1/\eps$ space whereas for a random sample one requires $1/\eps^2$ space. Full details are in the paper and its related literature.} or even precise answers as in the simple cases listed above.

\subsection{Metrics and Their Uses}
\label{sec:metrics}

Tracking these metrics provides more than just the ability to monitor simple statistics. It allows for more sophisticated use cases such as anomaly detection and drift detection. For anomaly detection, if we are monitoring a numeric variable that follows a Gaussian distribution, by tracking the mean and the standard deviation we can now detect anomalies by highlighting individual values that are too many standard deviations away from the mean. Of course there are many other techniques for detecting anomalies: some can be done based on information available from the metrics we store, and others must use a sample or connect to the stream using its own sketch. Random cut forest \cite{guha2016robust} is an example anomaly detection algorithm that maintains its own state, where the \emph{finalize} operation requires a data point as input and provides an anomaly score for it using the internal state.

For drift detection, we follow the observations in \cite{breck2019data}, and aim to have tests that (1) non-experts can intuitively tune and understand, and (2) detect not only the probability of two distributions being different, but also measure their distance. To elaborate on (2), classic methods for drift detection aim to answer ``how likely is collection $A$ to be drawn from the same distribution as collection $B$?'' 
\cite{breck2019data} describe a system where all models process vast amounts of data; in such a scenario the statistical aspect is not necessary and it suffices to answer ``Is the distance between the distribution of collection $A$ and that of collection $B$ larger than a threshold $\eps$?''
Since our system is designed for sparse loads as well, we do not remove the statistical component and aim to answer a question combining both aspects:
``how likely is collection $A$ to be drawn from a distribution that is $\eps$ close to distribution $B$?''

The tests we chose (other than the custom option) are variants of the student $t$-test and the two-sample Kolmogorov-Smirnov test. The student $t$-test essentially aims to discover whether the mean of two distributions are the same. Given a threshold $\eps$, our test computes the likelihood of the two means not being $\eps$ away from each other.
Tuning this $\eps$ is quite intuitive since it relates to the mean value, which is an easy concept when compared to metrics such as KL-divergence or cosine similarity. The metrics required to perform this test are the means, standard deviations, and the number of observations from each distribution.

The two-sample Kolmogorov-Smirnov test is a non-parametric test measuring whether two distributions are the same. Its distance variant is known as the $L_\infty$ distance of the distributions defined as $\|p-q\|_\infty = \max_x |p(x) - q(x)|$. Here, $p$ and $q$ are distributions and $p(x)$ is the probability of observing item $x$. When plotting a histogram for both distributions, this quantity is the bar in which $p$ and $q$ have the farthest value. This intuitive explanation makes it easy to set the threshold distance. In order to estimate this distance we require an (approximate) histogram of the empirical distributions, and the number of observed items.

For numerical inputs, both tests are applicable. For categorical inputs there is no concept such as a mean value, but the $L_\infty$ concept still applies. 

\subsection{Application for Detecting Drift in Model Fairness and Bias Statistics}
\label{sec:bias_drift}

Algorithmic bias, discrimination, fairness, and related topics have been studied across disciplines such as law, policy, and computer science \cite{barocas2016big}. An automated decision making model may be considered biased if it discriminates against certain features or groups of features. The ML models powering these applications learn from data and this data may reflect disparities or other inherent biases. For example, the training data may not have sufficient representation of various feature groups or may contain biased labels.

There are several metrics to measure bias corresponding to different notions of fairness \cite{das2021fairness}. Even considering simple notions of fairness leads to many different measures applicable in various contexts (e.g., regulations, laws, ethical principles, and others). Consider fairness with respect to gender or age, for example, and, for simplicity, 
that there are two relevant classes of roughly the same size.
In the case of an ML model for lending, we may want small business loans to be issued to equal numbers of both classes. Or, when processing job applicants, we may want to see equal numbers of members of each class hired.

Measuring the initial bias of our ML model is just a first step, but is not sufficient to guarantee that the model will remain fair over time. In fact, our model might become biased due to a shift in the distribution of the input data, and without a proper monitoring system this might evolve in producing biased automated predictions.
To tackle this issue, 
\modelmonitor offers a large variety of different fairness metrics to monitor -- for a comprehensive list of the implemented metrics see \cite{das2021fairness}) -- including Equal Opportunity \cite{hardt2016equality} and Statistical Parity.

\subsection{Detecting drift in model feature attributions}
\label{sec:feature_attr_drift}

Feature attribution methods such as LIME~\cite{ribeiro2016should} and SHAP~\cite{lundberg2017unified} provide importance scores for the individual input features, that is, how important a feature is to the predictions of the model. These scores can be computed at the instance level or at a global dataset level. A change in feature attribution scores can be indicative of a change in the underlying data, and consequently, the accuracy of the model on the new data~\cite{google_drift}. A drift in feature attribution can provide an early signal of potential degradation in model accuracy without the need to collect ground truth labels, which is a time-consuming and expensive task.

Feature attribution drift detection in \modelmonitor builds on top of the feature attribution functionality in SageMaker Clarify~\cite{hardt2021amazon}. Using Clarify, users can compute SHAP feature attributions by specifying the inputs, the address of the Endpoint where the model is deployed, and SHAP algorithm parameters~\cite{clarify_shap_config}. \modelmonitor additionally provides the capability to detect changes in attributions over time by capturing the model inputs / outputs over a specified period of time.
We can detect the drift by comparing how the ranking of the individual features changed from training data to inference data. In addition to being sensitive to changes in ranking \textit{order only}, we also want to be sensitive to the raw attribution scores of the features. For instance, given two features that fall in the ranking by the same number of positions going from training to inference data, we want to be more sensitive to the feature that had a higher attribution score in the training data. With these properties in mind, we use the Normalized Discounted Cumulative Gain (NDCG) score for comparing the feature attribution rankings of training and inference data.

Specifically, let $F = [f_1, \ldots, f_m]$ be the list of features sorted w.r.t. their attribution scores in the training data. Let $a(f)$ denote the attribution score of feature $f$ on the training data. 
Finally, let $F' = [f'_1, \ldots, f'_m]$ be the list of features sorted w.r.t. their attribution scores in the inference data.
Then, we can compute the NDCG = $\frac{DCG}{iDCG}$ where $DCG = \sum_{i=1}^{m} \frac{a(f'_i)}{log_2(i+1)}$ and $iDCG = \sum_{i=1}^{m} \frac{a(f_i)}{log_2(i+1)}$.

The quantity DCG measures the extent to which features with high attribution in the training data are also ranked higher in the feature attribution computed on the inference data. The quantity iDCG measures the ``ideal score'' and is just a normalizing factor to ensure that the final quantity resides in the range [0, 1], with 1 being the best possible value. An NDCG value of 1 means that the feature attribution ranking in the inference data is the same as the one in the training data. We automatically raise an alert in \modelmonitor if the NDCG value drops below 0.9.

\section{Quantitative Evaluation}\label{sec:quantitative}
\subsection{Value of Adaptive Retraining} \label{sec:adaptive}
A common scenario brought up by users is the need to retrain the model. It is a common scenario that in order to maintain model performance, one retrains the model at regular and frequent time intervals. This retraining procedure can be costly as it requires compute resources for training, as well as an overhead paid for a gradual deployment replacing the old model with the new one. Users wish to monitor the quality of the deployed model and retrain it only when needed. The actual value of this varies for different scenarios, we will now show an experiment evaluating it in a synthetic controlled environment. 

\begin{figure}
\centering
    \includegraphics[width=0.8\textwidth]{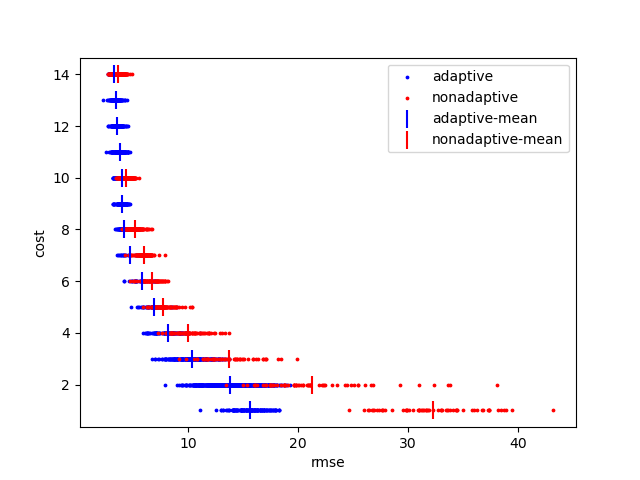}
 \caption{Comparison of {\bf adaptive} and {\bf nonadaptive} methods, plotting their RMSE loss vs.\ cost incurred. The dots represent individual experiments. The bars represent the average RMSE over all points with the same cost for a given method.}
    \label{fig:adaptive}
\end{figure}

The exact details of the environment are given in Appendix~\ref{app:adaptive_training}. The high level idea for the environment is that there is a model that drifts over time, and the system can choose when to retrain in order to better track it. We measure for an adaptive system that retrains based on error and a nonadaptive system that retrains at regular intervals, their cost (number of retrains) vs.\ error (RMSE). In Figure~\ref{fig:adaptive} we show a scatter plot of the systems in repeated experiments with different random seeds; the bars represent the mean RMSE over all experiments with the same cost. It is clear that the adaptive system provides a significantly better tradeoff.  

\subsection{Data Collection Overhead}

\modelmonitor is designed to allow monitoring with little to no impact on production workloads. A key aspect of this design was to perform data collection independently from analysis. Our solution, described in \ref{data_collection}, involves collecting data to disk and asynchronously uploading to S3 for later analysis.

When comparing SageMaker Endpoints with data capture enabled and those with it disabled, in a variety of hardware configurations (that is, different EC2 instance types), we were only able to detect minimal (smaller than 1\%) changes in CPU and memory. 

\section{Case Study, Use Cases, and Deployment Insights}\label{use-cases}
\begin{figure*}
     \centering
     \begin{subfigure}[b]{0.6\textwidth}
         \centering
         \includegraphics[width=\textwidth]{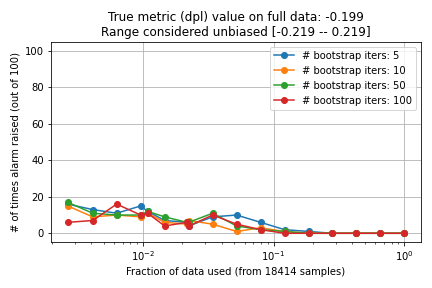}
         \caption{No bias: $C=[-0.219, 0.219]$}
     \end{subfigure}
     \\
     \begin{subfigure}[b]{0.6\textwidth}
         \centering
        \includegraphics[width=\textwidth]{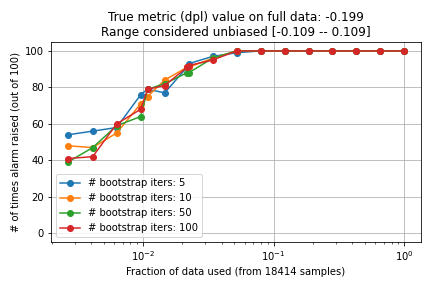}
         \caption{Medium bias: $C=[-0.109, 0.109]$}
     \end{subfigure}
     \\
     \begin{subfigure}[b]{0.6\textwidth}
         \centering
         \includegraphics[width=\textwidth]{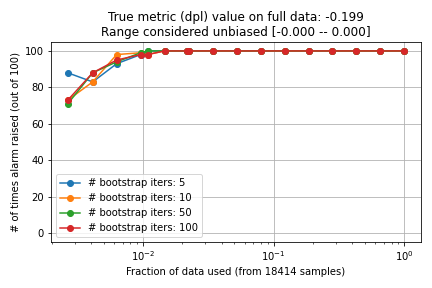}
         \caption{High bias: $C=[0,0]$}
     \end{subfigure}
    \caption{UCI Adult dataset with DPL as the bias metric. The true value of the bias metric on the full data is $b=-0.199$. 
    The three plots show the fraction of alarms raised under different acceptable ranges ($C$) of the observed bias metric. If the actual values $b$ of the bias metric is contained within $C$, no alarms should be raised.
    }
    \label{fig:dpl}
\end{figure*}

Next, we present a  case-study of using \modelmonitor for detecting drift in bias metrics, some common use-cases, and insights learned from deployment in the real-world.

\subsection{Case Study}

\noindent\textbf{Goal.} The goal of this case study is to investigate a concrete scenario involving bias drift detection with \modelmonitor, evaluate different hyperparameter choices, and select the most promising ones. Similar kind of evaluation strategies can be used for other drift detection tasks.

\noindent\textbf{Setup.}
For the case study, we use the Adult dataset~\cite{Dua:2019} from the UCI Machine Learning repository. The datasets was split into train and test partitions of size $60\%$ and $40\%$ respectively. We train a Logistic Regression classifier from scikit-learn%
\footnote{\url{https://scikit-learn.org}}
on the data and note the value of the bias metric (denoted as $b$) on the full test set. The 
bias metric that we report on here is the Difference in Positive Proportions in Observed Labels (DPL)~\cite{das2021fairness}. The takeaways are similar for other metrics like Accuracy Difference, Difference in Positive Proportions in Predicted Labels, and difference in FPR and FNR for different groups.

Next, to evaluate the accuracy of different hyperparameter configurations for detecting a bias drift, we simulate the cases of \textit{low bias}, \textit{medium bias}, and \textit{high bias} in the observed subset that \modelmonitor will see in deployment. No bias is simulated by specifying an acceptable range of the bias metric that already contains the actual metric value $b$ on the full test set ($b=-0.199$ in this case). Specifically, the range is calculated as $C = [1.1b, -1.1b]$. Since the actual value is contained within the acceptable range, we expect very few to no alarms to be raised. Medium and high bias are simulated by specifying the ranges to be $C=[0.55b, -0.55b]$ and $C=[0, 0]$, respectively.

The hyperparameters in this case are the bootstrap sample size, and the number of bootstrap iterations ($n_{boot}$). For each bootstrap iteration, we
sampled $80\%$ of the available sample with replacement. For each value of $C$ and $n_{boot}$, we run the bootstrap-based alarm raising procedure on
different sample sizes. For each sample size, we run the alarm raising procedure $100$ times and report how many times an alarm was raised.

\noindent\textbf{Takeaways.}
The results for all three cases of bias are shown in Figure~\ref{fig:dpl}. The results show that:
\begin{enumerate}
    \item In all cases, as expected, as the sample size for building the bootstrap confidence intervals increases, the accuracy of the alarm raising procedure increases. Specifically, in all three cases (low, medium, high bias), once $10\%$ ($\sim$2,000 samples) of the data is observed, the alarm raising procedure is almost always $100\%$ accurate -- that is, $0$ alarms in the case of no bias and $100$ alarms in the case of medium and high bias.
    \item The only exception to the takeaways above is the case of using $5$ bootstrap iterations in the case of no bias. As the number of bootstrap iterations increases, the variance estimates and consequently the accuracy is expected to improve.
    \item The case of high bias requires much fewer samples to accurately raise alarms, which is due to the fact that the actual value of the metric ($b=-0.199$) has a large difference from the acceptable range $C=[0,0]$.
    \item Even with $1\%$ ($\sim$200 samples) of the samples available, we achieve a reasonably high accuracy ($>80\%$ in high or no bias cases in Figure~\ref{fig:dpl}, slightly less in the case of medium bias).
\end{enumerate}

Given that ~200 samples were often sufficient to achieve high accuracy, we set the bootstrap sample size to 200.
Similarly, since the number of bootstrap iterations showed little impact on the accuracy of the alarm raising procedure, we set it to $5$.

\subsection{Examples of Use Cases}
We next highlight two common use cases with \modelmonitor.

\textbf{Market Intelligence.}
Market intelligence models operate on a large variety of inputs from disparate sources (e.g., first party, third party, and survey data) and are used to more effectively cater to their target audience. The applications of these models extend into industries such as automotive, insurance, media consumption, finance, and travel. The performance of these models can suffer from data drifts (e.g., change in the statistics of input variables) as well as concept drift (e.g., change in relationships between different variables).

To this end,
\modelmonitor uses rules such as \textit{data\_type\_check, completeness\_check, baseline\_drift\_check, missing\_column\_check, extra\_column\_check}, and \textit{categorical\_values\_check} to detect data drift. For instance, consider an ML model used by a financial institution to predict loan approval for an applicant. When training this model, age is used as a feature and is always present. However, later on, the financial institution decided to make age optional when applying for loans. This change caused the training data and the real world input data to be different. In this case, \modelmonitor detected this change in the real world data and the model maintainers were alerted of the violation.\footnote{\scriptsize \modelmonitor violations: \url{https://docs.aws.amazon.com/sagemaker/latest/dg/model-monitor-interpreting-violations.html}}

Model quality monitoring\footnote{\scriptsize Model quality monitoring: \url{https://docs.aws.amazon.com/sagemaker/latest/dg/model-monitor-model-quality.html}} monitors the 
performance of a model by comparing the predictions that the model makes with the actual ground truth labels that the model attempts
to predict. 
Model quality monitoring jobs compute different metrics\footnote{\scriptsize \modelmonitor quality metrics: \url{https://docs.aws.amazon.com/sagemaker/latest/dg/model-monitor-model-quality-metrics.html}} depending on the ML problem type. 
These include model accuracy, confusion matrix, recall, area under the curve (AUC), F1 score, and F2 score. When a certain metric shifts beyond 
or below a threshold value -- determined by the baselining job -- a violation is triggered allowing customers to track the 
model performance against the real world data. 
For instance, consider the loan approval prediction model as before. 
According to Freddie Mac,\footnote{\scriptsize Historical mortgage interest rates: \url{http://www.freddiemac.com/pmms/}} mortgage interest rates have been steadily dropping since November 2018. All other things being equal, as interest rates drop, the loan amount approved for a given income is higher because monthly payments are lower. A model trained using a dataset from a period of higher interest would make more loan rejection predictions for a given income and loan amount compared to real life data. In this case, \modelmonitor detected this change in the real world data and the model maintainers were alerted of the diverging F score.

\textbf{Detecting data drift in NLP models.}
NLP encoders like word\-2vec~\cite{mikolov2013distributed}, BERT~\cite{devlin2018bert} and RoBERTa~\cite{liu2019roberta} are used in a wide array of applications, ranging from chat bots and virtual assistants, to machine translation and text summarization. 
These encoders operate by converting input words or sequences of words into 
word-level embeddings. These embeddings are then used by downstream task-specific models. A change in the distribution of the input text can clearly impact the performance of the downstream model.

\begin{figure}
\centering
  \includegraphics[width=12cm]{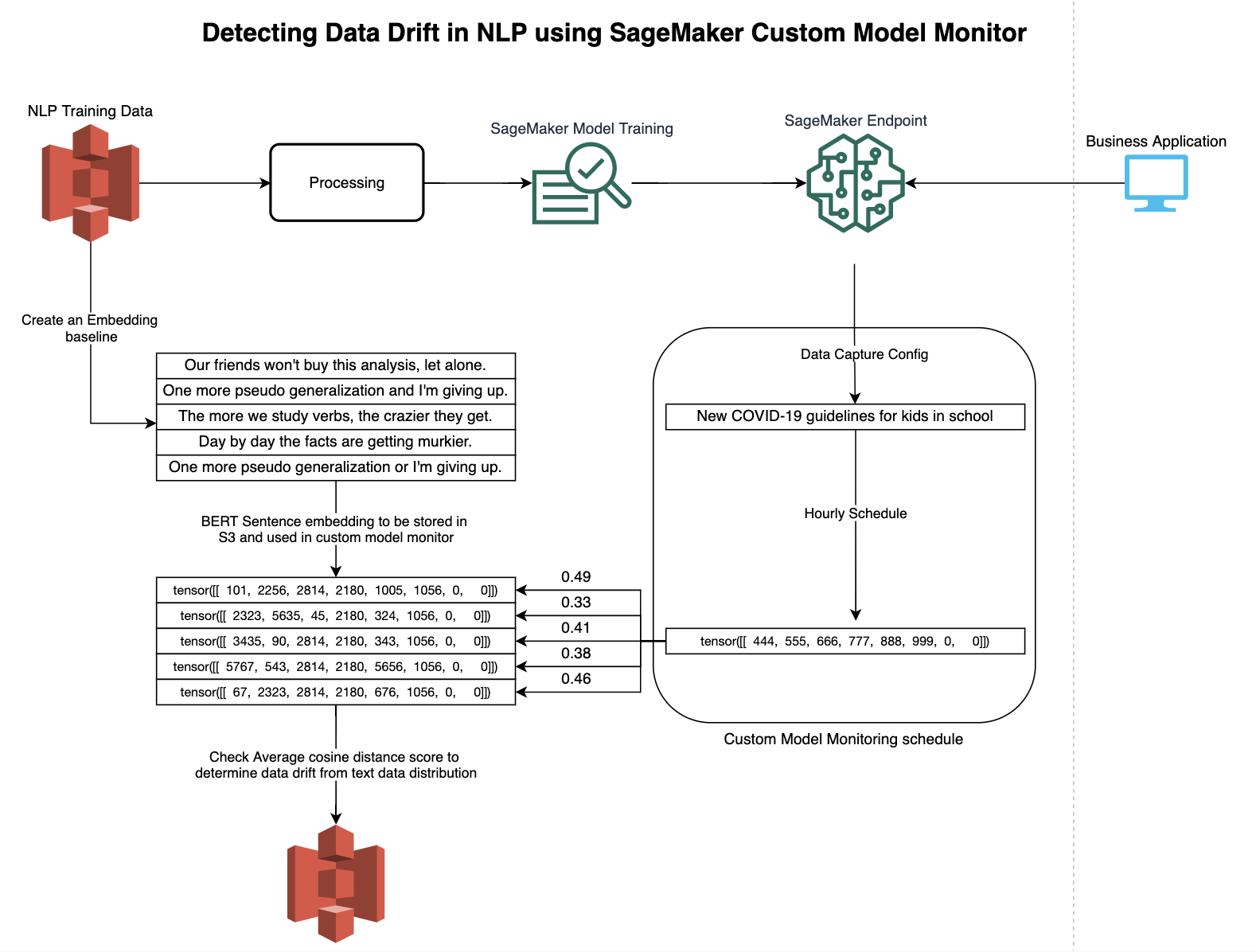}
  \caption{Detecting data drift in NLP models using custom monitoring schedule in \modelmonitor}
  \label{fig:nlp-data-drift-bert}
\end{figure}

However, due to specific structure of text, monitoring text data is a challenging task.  Unlike tabular data which is often fixed-dimensional and bounded, text data is often free form. To overcome these issues, we decided to use \modelmonitor on the embeddings of the text data as opposed to the raw text itself.
Figure~\ref{fig:nlp-data-drift-bert} shows an example of configuring a custom monitoring schedule to detect drifts in text data.
In this use case,\footnote{\scriptsize \url{https://aws.amazon.com/blogs/machine-learning/detect-nlp-data-drift-using-custom-amazon-sagemaker-model-monitor}} we showcase how we detect the data drift in a \textit{Bring Your Own Container} (BYOC) fashion.\footnote{\scriptsize BYOC with \modelmonitor: \url{https://docs.aws.amazon.com/sagemaker/latest/dg/model-monitor-byoc-containers.html}}

In this example, we start from a BERT encoder and add task-specific layer(s) on top. Given the training dataset for the target task, in addition to training the task specific layer(s), the BERT encoder can optionally be fine-tuned as well~\cite{devlin2018bert,sun2019fine}.
Given an input sequence of length $L$, the encoder results in an embedding matrix  $E \in \mathbb{R}^{L \times D}$ where $D$ is the size of individual word embedding vector. For each input sequence, we average the word level embeddings to get a single sequence level embedding vector of size $D$.
Given an input training set of size $N$, we store the per-sequence embedding vectors for all the individual texts $\{e_i\}_{i=1}^{N}$ in a S3 bucket.

For detecting drift at deployment time,\footnote{\scriptsize Evaluation script to measure data drift using cosine similarity: \url{https://github.com/aws-samples/detecting-data-drift-in-nlp-using-amazon-sagemaker-custom-model-monitor/blob/main/docker/evaluation.py}}
we adopt the following procedure. Given an input text $t$, we compute its sequence level embedding $e_t$, and compute the average pairwise cosine similarity score between $e_t$ and each of the training set embeddings $\{e_i\}_{i=1}^{N}$. Then, we aggregate across multiple inputs over the monitoring window (e.g., by averaging the scores), and consider the input distribution to have drifted if the aggregated score falls below a chosen threshold.
Using \modelmonitor, we setup a monitoring schedule that runs every hour to continuously look for data drift. In case a violation is detected, we send CloudWatch alerts and SNS notifications to allow the customers to re-train the model.

\subsection{Deployment Insights and Lessons Learned in Practice}\label{lessons-learned}
As model monitoring is important and broadly applicable to any production deployment of ML based systems, \modelmonitor is used by customers worldwide from different industries including financial services, automotive, software, telecommunications, and retail.

\textbf{Improving getting started}: We launched \modelmonitor primarily with programmatic interfaces to setup monitoring. From interviews with customers post-launch, we noticed about one sixth of interviewees preferred using a graphical interface for setting up monitoring, and feature parity with programmatic interfaces was important. Based on this feedback, we created a graphical guided process that enables monitoring and hides most options by picking sensible defaults, while still allowing users to override these when needed.

\textbf{Ease of use beyond tabular data}: We focused on providing a default no-code experience for tabular ML use cases. However, customers have a variety of use cases beyond tabular models, such as monitoring image classification models or metrics associated with NLP models. While \modelmonitor was designed to be extensible by bringing an analysis container, only one in three customers choose to go this route. These customers choose solutions like \modelmonitor to apply research into turnkey experiences so they don't have to. By leveraging our support for custom Docker images, we created documentation and blog posts showcasing how customers can use \modelmonitor even for these use cases.

\textbf{Hooks for augmenting data - pre-/post- processing hooks}: Our initial implementation performed model monitor analysis directly on the data as captured by the Endpoint. From discussions with prospective customers, many of them send data in production that is not amenable to direct analysis: inference payload may be sent compressed into the Endpoint, or without transformations like one-hot encoding used by the actual ML model. Additionally, the format our results are stored in may need to be tweaked. In response, we added pre- and post-processing hooks to enable performing custom logic before and after the analysis. Today, almost half of our customers rely on this feature to augment the input and output data into the forms they need.

\textbf{Bias and feature attribution drift}: Our initial offering (launched in December, 2019) focused on monitoring of data drift. We learned from our customers that they would like to not only monitor the quality of the deployed model as a whole but also understand whether there are emerging differences in the predictive behavior of the model across different groups. Although the initial data or model may not have been biased, changes in real-world conditions may cause bias to develop over time in a model that has already been trained. For example, a substantial change in home buyer demographics could cause a home loan application model to become biased if certain populations were not present in the original training data. To address such challenges, we expanded the functionality of \modelmonitor by integrating with SageMaker Clarify (in December, 2020), thereby enabling model owners to monitor for drift in bias and feature attributions, and take remedial actions based on the alerts.

\section{Related Work}\label{sec:related}
Detecting changes in the features or model outputs and verifying the validity of model inputs is a longstanding problem in ML. Here, we give a background on the techniques as well as tools available that address this problem.

Some of the simplest techniques to verify feature validity involve adding user-defined tests. For instance, \cite{schelter2018automating} suggests adding tests to verify the minimum / maximum of input features to the model. More complex tests like comparing the histogram of the feature can also be added. On the plus side, these tests are easily interpretable due to their simplicity. However, they may require large configuration overhead from the user.

Drift detection can be posed as a null hypothesis significance testing (NHST) problem: Given two sets of samples, check via a test statistic (e.g., mean) if the two samples come from the same distribution. Some of the most well-known tests are Student's t-test and Kolmogorov–Smirnov test~\cite{wasserman2004all,murphy2012machine}. For higher dimensional data, more advanced tests, for instance, Maximum Mean Discrepancy can also be used~\cite{gretton2012kernel}. However, these tests require more fine-tuning (e.g., selecting the kernel and its hyperparameters). Confidence intervals offer an alternative to performing NHST. 
In \modelmonitor, we leverage bootstrap based confidence intervals for detecting drifts~\cite{efron1994introduction}.

Several open source and commercial tools provide users with the ability to monitor predictions of deployed ML models, e.g., Deequ \cite{schelter2018automating}, Evidently \cite{evidently}, Fiddler's  Explainable Monitoring \cite{fiddler_monitor}, Google Vertex AI Model Monitoring \cite{google_drift}, IBM Watson OpenScale \cite{ibm_monitor}, Microsoft Azure MLOps \cite{azure_monitor}, and Uber's Michelangelo platform~\cite{uber_michelangelo}.


\modelmonitor detects potential  performance degradation in a model-agnostic manner and without needing to store the full training dataset. However, specialized solutions that leverage the model internals and the training data can be used to quantify the extent of drift and take targeted corrective actions (e.g.,~\cite{NEURIPS2020_219e0524,lipton2018detecting,reddi2015doubly,wu2019domain}). Due to their model and data introspective nature, these solutions are beyond the scope of this paper.

\section{Conclusion}\label{sec:conclusion}

Motivated by the need for maintaining high quality machine learning models once they are deployed, we presented Amazon SageMaker \modelmonitor/, a system for automatically detecting data, concept, bias, and feature attribution drift in models in real-time and providing alerts so that model owners can take remedial actions. We described the key design considerations and technical architecture of \modelmonitor along three dimensions: data collection, data analysis, and job scheduling. For each of these, we highlighted the technical challenges we faced, along with how we addressed them within our solution. We discussed the methodology for data drift detection and application for detecting drift in model bias metrics and feature attributions, a case study, common use cases, and deployment insights. As our system is a scalable, cloud-based, fully managed model monitoring service designed to cater to the needs of customers from different industries, the experience and insights from our work are likely to be useful for researchers and practitioners working in the emerging field of machine learning operations (MLOps).
Guaranteeing a model's performance after deployment is an area of natural focus as machine learning systems make their way into critical industries, and monitoring models in production is a key part of the solution.

\section*{Acknowledgments}
The authors would like to thank other members of the Amazon AWS AI team for their collaboration during the development and deployment of Amazon SageMaker Model Monitor, and 
Jay Casteel,
Xiaoguang Chen,
Urvashi Chowdhary,
Venkatesh Krishnan,
Maximiliano Maccanti,
Jason McMaster,
Shikher Mishra,
Arun Nagarajan,
Andrea Olgiati,
Archana Padmasenan,
Aakash Pydi,
Rakesh Ramakrishnan,
Dayanand Rangegowda,
Sanjiv Das,
James Sanders,
Shruti Sharma,
Rohit Singh,
Abhishek Taneja,
and
Pinar Yilmaz
for insightful feedback and discussions.

\bibliography{bibfile}
\bibliographystyle{acm}

\clearpage
\appendix

\section{Enabling production monitoring}\label{app:enabling}
As we describe above, we placed particular focus on making model monitoring as close to a no-code experience as possible. Users should not need to change how their model is built: we achieved this by decoupling data collection from analysis. The code to enable data collection, when using the SageMaker Python SDK, is as follows:

\begin{scriptsize}
\begin{lstlisting}[language=Python]
from sagemaker.model_monitor import DataCaptureConfig

data_capture_config = DataCaptureConfig(
    enable_capture=True,
    sampling_percentage=100,
    destination_s3_uri='s3://path/for/data/capture'
)

predictor = model.deploy(
    initial_instance_count=1,
    instance_type='ml.m4.xlarge',
    data_capture_config=data_capture_config
)
\end{lstlisting}
\end{scriptsize}

After that is done, to enable model monitor it suffices to describe where the baseline dataset is stored, as well as the desired hardware configuration to use. 

\begin{scriptsize}
\begin{lstlisting}[language=Python]
from sagemaker.model_monitor import DefaultModelMonitor
import sagemaker.model_monitor.dataset_format as datafmt

my_monitor = DefaultModelMonitor(
    role=role,
    instance_count=1,
    instance_type='ml.m5.xlarge',
    volume_size_in_gb=20,
    max_runtime_in_seconds=3600,
)

my_monitor.suggest_baseline(
    baseline_dataset='s3://path/validation-dataset.csv',
    dataset_format=datafmt.DatasetFormat.csv(header=True),
)
\end{lstlisting}
\end{scriptsize}

Lastly, once baseline job has run, (in this case, hourly) monitoring can be enabled with:

\begin{scriptsize}
\begin{lstlisting}[language=Python]
from sagemaker.model_monitor import CronExpressionGenerator

hourly_cron = CronExpressionGenerator.hourly()

my_monitor.create_monitoring_schedule(
    monitor_schedule_name='my-monitoring-schedule',
    endpoint_input=predictor.endpoint_name,
    statistics=my_monitor.baseline_statistics(),
    constraints=my_monitor.suggested_constraints(),
    schedule_cron_expression=hourly_cron,
)
\end{lstlisting}
\end{scriptsize}

Interested readers can refer to our full documentation at \url{https://sagemaker.readthedocs.io/en/stable/amazon_sagemaker_model_monitoring.html}

\section{Sample monitoring job results}\label{app:samplereport}


We will pick regression model quality reports to illustrate here. We support standard regression metrics out of the box, and compute standard deviations by following the bootstrapping approach described in \ref{use-cases}

\begin{scriptsize}
\begin{verbatim}
{
  "version" : 0.0,
  "dataset" : {
    "item_count" : 4,
    "start_time" : "2020-10-29T22:00:00Z",
    "end_time" : "2020-10-30T00:00:00Z",
    "evaluation_time" : "2021-10-13T00:17:07.894Z"
  },
  "regression_metrics" : {
    "mae" : {
      "value" : 0.5,
      "standard_deviation" : 0.2260776661041756
    },
    "mse" : {
      "value" : 0.5000000000000001,
      "standard_deviation" : 0.2260776661041756
    },
    "rmse" : {
      "value" : 0.7071067811865476,
      "standard_deviation" : 0.13949944301704126
    },
    "r2" : {
      "value" : -1.6666666666666674,
      "standard_deviation" : "NaN"
    }
  }
}
\end{verbatim}
\end{scriptsize}

The constraints for such data might be constructed as:

\begin{scriptsize}
\begin{verbatim}
{
  "version" : 0.0,
  "regression_constraints" : {
    "mae" : {
      "threshold" : 0.5,
      "comparison_operator" : "GreaterThanThreshold"
    },
    "mse" : {
      "threshold" : 0.5000000000000001,
      "comparison_operator" : "GreaterThanThreshold"
    },
    "rmse" : {
      "threshold" : 0.7071067811865476,
      "comparison_operator" : "GreaterThanThreshold"
    },
    "r2" : {
      "threshold" : -1.6666666666666674,
      "comparison_operator" : "LessThanThreshold"
    }
  }
}
\end{verbatim}
\end{scriptsize}

Note that our system is designed to throw alerts if thresholds are breached when taking the standard deviation into account (e.g., for GreaterThan thresholds, only if \(value > baseline + stddev\))

\section{Sample \modelmonitor Visualizations}
The outputs of \modelmonitor can be visualized through SageMaker Studio and CloudWatch Metrics. 

\textbf{SageMaker Studio} gives the user complete access, control, and visibility into each step required to build, train, and deploy models. Users can quickly upload data, create new notebooks, train and tune models, move back and forth between steps to adjust experiments, compare results, and deploy models to production all in one place. All ML development activities including notebooks, experiment management, automatic model creation, debugging, and model and data drift detection can be performed within SageMaker Studio.

\begin{figure}
\centering
  \includegraphics[width=8.5cm]{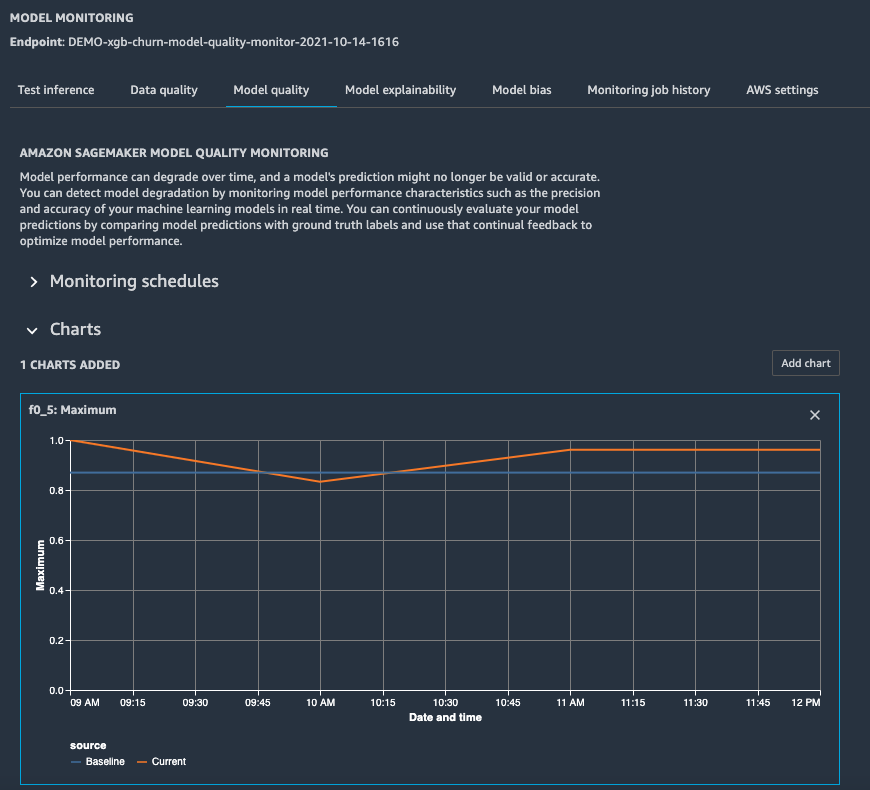}
  \caption{Model Quality tab in SageMaker Studio showing of \modelmonitor Model Quality calculation of the F0.5 score for a binary classification model}
  \label{fig:studiosample}
\end{figure}

To see the \modelmonitor results, users can navigate to the endpoint for which \modelmonitor is enabled and then select the type of monitoring results they want to view. Uesrs can customize and add different charts to match the metrics they want to visualize. A screenshot of this view is shown in \ref{fig:studiosample}.

\begin{figure}
\centering
  \includegraphics[width=8.5cm]{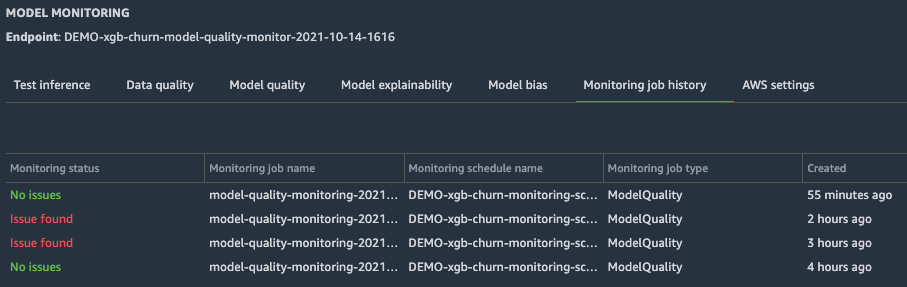}
  \caption{The ``Monitoring Job History'' tab in SageMaker Studio showing a list of \modelmonitor jobs that have run for this model along with a summary of the jobs.}
  \label{fig:studiojoblistsample}
\end{figure}

An overview of all the jobs, their statuses, and the number of issues identified can be found in the ``Monitoring Job History'' tab. A screenshot of this view is shown in \ref{fig:studiojoblistsample}.

\textbf{CloudWatch} provides the users with data and actionable insights to monitor applications, respond to system-wide performance changes, optimize resource utilization, and get a unified view of operational health. CloudWatch collects monitoring and operational data in the form of logs, metrics, and events, providing users with a unified view of AWS resources, applications, and services that run on AWS and on-premises servers. Users can use CloudWatch to detect anomalous behavior in your environments, set alarms, visualize logs and metrics side by side, take automated actions, troubleshoot issues, and discover insights to keep  applications
running smoothly.

\begin{figure}
\centering
  \includegraphics[width=8.5cm]{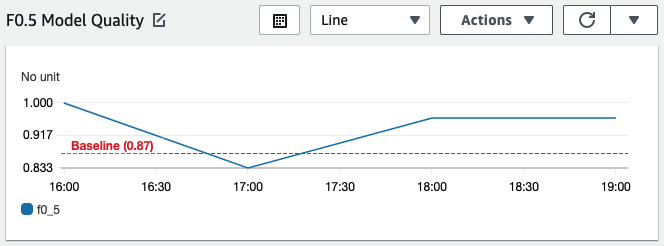}
  \caption{A chart showing how the F0.5 score changes for a binary classification model computed by \modelmonitor}
  \label{fig:cloudwatchsample}
\end{figure}

\modelmonitor can send metrics to CloudWatch Metrics as well. Dashboards containing different graphs can be created to monitor the metrics emitted by \modelmonitor. A sample chart is shown in \ref{fig:cloudwatchsample}. In addition, CloudWatch Metrics can be configured to send a notification if metrics drop below a defined threshold. Notifications generated by CloudWatch Metrics can be used to start a retraining event.

For example, one might be interested in using \modelmonitor integration with Amazon SageMaker Clarify in order to improve visibility into potential bias drift. For illustration we selected German Credit Data dataset from UCI \cite{Dua:2019}. This dataset contains 1000 labeled credit applications and some of features include information concerning credit purpose, credit amount, housing status, and employment history. For practical purpose, categorical attributes are converted to indicator variables, e.g., A14 means “no checking account”. We mapped the labels (Class1Good2Bad) to 0 for poor credit and 1 for good. As we discussed in Section \ref{sec:bias_drift}, although the initial data or model may not have been biased, changes in the world may cause bias to develop over time in a model that has already been trained. 

\begin{figure}
\centering
  \includegraphics[width=8.5cm]{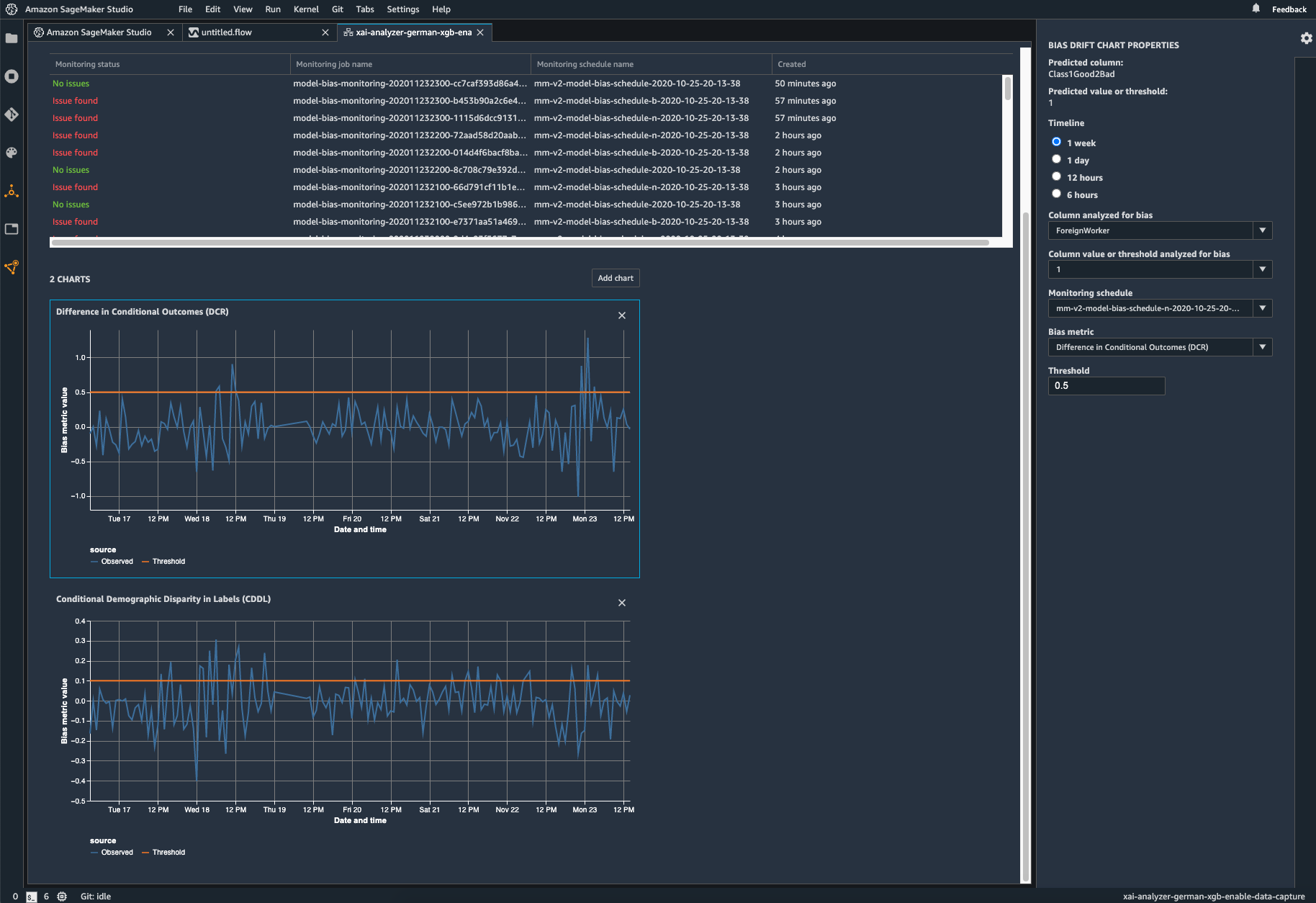}
  \caption{\modelmonitor controlling two bias metrics (DCR and CDDL \cite{das2021fairness}) with the feature “ForeignWorker” as sensitive attribute for German Credit Data dataset. Warnings are called when the value of the bias metrics is higher than the fixed threshold (horizontal orange lines in the monitoring plots).}
  \label{fig:german_bias_monitor}
\end{figure}

Figure \ref{fig:german_bias_monitor} shows \modelmonitor monitoring two bias metrics (i.e., Difference in conditional rejection (DCR) and Conditional Demographic Disparity (CDDL) -- see \cite{das2021fairness} for more details) considering the feature “ForeignWorker” as sensitive attribute to protect and monitor. The system rises issues if the value of the bias metrics is higher than their fixed thresholds -- in our example 0.5 for DCR, and 0.1 for CDDL.

\begin{figure}
\centering
  \includegraphics[width=8.5cm]{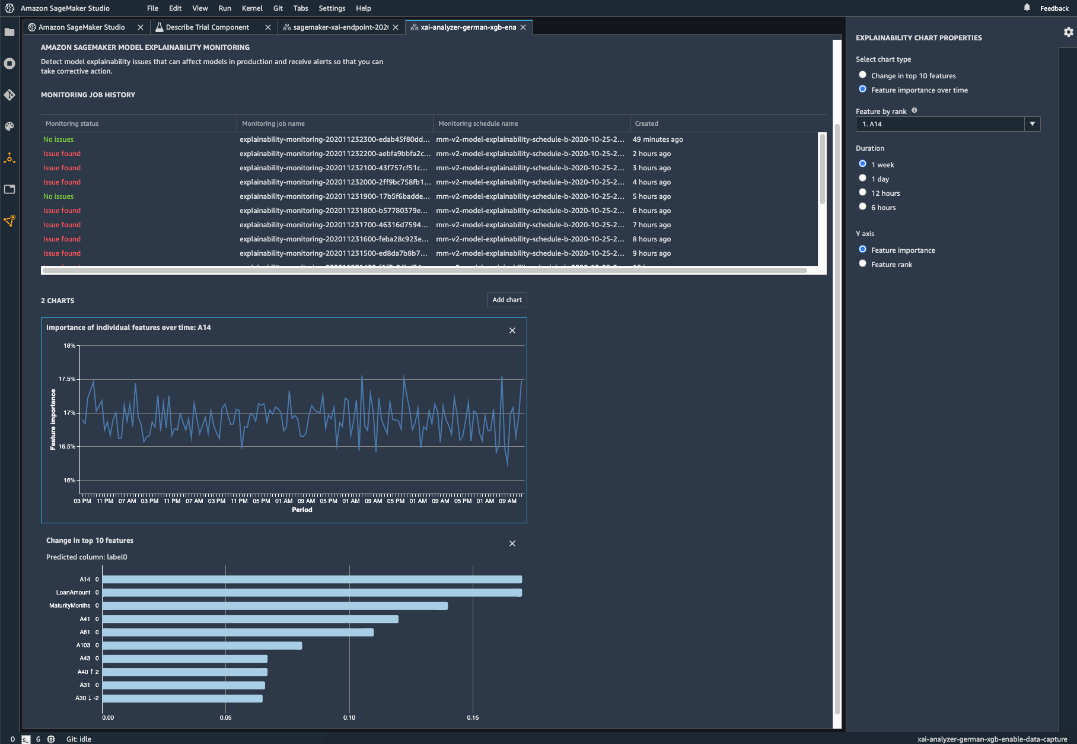}
  \caption{\modelmonitor analyzing the feature importance for German Credit Data dataset. Feature A14 (“no checking account”) is the most important one and \modelmonitor is able to analyze how the importance of feature A14 evolves through time.}
  \label{fig:german_feature_importance}
\end{figure}

In Figure \ref{fig:german_feature_importance} we can see the monitoring of the feature importance for German Credit Data dataset. We can note that feature A14 (“no checking account”) is the most important one and we can use \modelmonitor to control if the importance of feature A14 remains similar through time, or there are changes in feature importance instead.

\section{Details for the experiment of Section~\ref{sec:adaptive}} \label{app:adaptive_training}

In the experiment of Section~\ref{sec:adaptive}, the model is a linear regression model with a single feature. The data it observes is i.i.d.\ Gaussian with mean zero and standard deviation one. The true labels are generated according to an linear function over the input $y = ax + b$ where $x$ is the input, $a,b$ are scalars and $y$ is the label. over time, the parameters of this generative model shift. We chose to the parameters, after every 100 examples, a Gaussian noise ${\cal N}(1,1)$ meaning having standard deviation 1 and mean 1. We chose a mean of 1 to make sure the model does not return to its previous state, as this is less frequent in high dimensional data. 
In our setup a system has a linear model that outputs predictions and we track the RMSE defined as the square root of the mean squared error. At any given point the system can retrain. By doing so, it updates the model to be the same as the generative model but pays a cost of 1. 

We compare two setups: (1) The first setup is called {\bf nonadaptive}, and we retrain the model at regular intervals. We control the cost, i.e., number of retrains by setting the length of the regular interval. By setting a large interval we pay a low cost but suffer a large RMSE, and conversely a small interval leads to a low RMSE but large cost; (2) The second setup is called {\bf adaptive}, and we retrain a model once its RMSE in the past 100 observations exceed some threshold. Here, this threshold is the knob we use to balance cost vs.\ RMSE. 

We repeat the following experiment with different random seeds. We draw at random which setup to use, {\bf adaptive} or {\bf nonadaptive}. We draw at random the parameter of the system, meaning the interval length for {\bf nonadaptive} and threshold for {\bf adaptive}. The model drift also behaves in an independent way in each round. Every such round provides a single triplet in the form (technique, cost, RMSE). Figure~\ref{fig:adaptive} shows a scatter plot for these points.

\end{document}